\documentclass[conference,onecolumn,12pt]{IEEEtran}
\usepackage{amsmath,amssymb,amsfonts,mathrsfs,mathtools,bm, bbm, dsfont,mathrsfs,amsthm,blkarray}
\usepackage[mathscr]{eucal}
\usepackage[dvipdfm]{graphicx}
\usepackage[dvipsnames]{xcolor}
\usepackage{makecell}
\usepackage{float}
\usepackage{caption}
\usepackage{xcolor}    

\definecolor{deepblue}{RGB}{23,5,157}
\definecolor{deepred}{RGB}{192,0,0}

\usepackage{epsfig,epsf,psfrag,latexsym, graphics}
\usepackage[ruled,vlined]{algorithm2e}

\usepackage{booktabs}
\usepackage{multirow,multicol}

\makeatletter
\renewenvironment{abstract}{%
  \if@twocolumn
    \section*{Abstract}%
  \else
    \begin{center}
      {\bfseries Abstract}
    \end{center}%
  \fi
}{}
\makeatother

\usepackage[breaklinks,colorlinks, linkcolor=MidnightBlue, anchorcolor=MidnightBlue, citecolor=MidnightBlue, urlcolor=MidnightBlue]{hyperref}
\usepackage[hyphenbreaks]{breakurl}
\usepackage{cite}
\usepackage{xurl}

 \def\old#1{}    

\usepackage{orcidlink}

\def\beq{\begin{equation}}
\def\eeq{\end{equation}}
\def\bea{\begin{eqnarray}}
\def\eea{\end{eqnarray}}
\def\ba{\begin{array}}
\def\ea{\end{array}}

\def\bitem{\begin{itemize}}
\def\eitem{\end{itemize}}
\def\ben{\begin{enumerate}}
\def\een{\end{enumerate}}

\def\ie{{\it i.e.,\ \/}}

\newcommand{\mbbE}{\mathbb{E}}

\newcommand{\Xmsc}{\mathscr{X}}

\def\epsilonbf{\hbox{\boldmath$\epsilon$\unboldmath}}

\def\thetabf{{\mbox{\boldmath$\theta$\unboldmath}}}

\def\ebf{{\bm e}}

\def\ubf{{\bm u}}
\def\vbf{{\bm v}}

\def\xbf{{\bm x}}

\def\xbf{{\bm x}}

\def\Vbf{{\bm V}}

\def\Xbf{{\bm X}}

\def\Zbf{{\bm Z}}

\def\Uc{{\cal U}}

\def\Xc{{\cal X}}

\newcommand{\beqa}{\begin{eqnarray}}
\newcommand{\eeqa}{\end{eqnarray}}
\newcommand{\beqan}{\begin{eqnarray*}}
\newcommand{\eeqan}{\end{eqnarray*}}

\newcounter{l1}
\newcounter{l2}
\newcounter{l3}
\newcommand{\bdotlist}{\begin{list}{$\bullet$}{}}
\newcommand{\bboxlist}{\begin{list}{$\Box$}{}}
\newcommand{\bbboxlist}{\begin{list}{\raisebox{.005in}{{\tiny
$\blacksquare$ \ \ }}}{}}
\newcommand{\bdashlist}{\begin{list}{$-$}{} }
\newcommand{\blist}{\begin{list}{}{} }
\newcommand{\barablist}{\begin{list}{\arabic{l1}}{\usecounter{l1}}}
\newcommand{\balphlist}{\begin{list}{(\alph{l2})}{\usecounter{l2}}}
\newcommand{\bAlphlist}{\begin{list}{\Alph{l2}.}{\usecounter{l2}}}
\newcommand{\bdiamlist}{\begin{list}{$\diamond$}{}}
\newcommand{\bromalist}{\begin{list}{(\roman{l3})}{\usecounter{l3}}}

\begin{document}
\include{pythonlisting}
\title{AI Foundation Model  for Time Series with Innovations Representation}
\author{\IEEEauthorblockN{Lang Tong}\\
\vspace{-1em}
\IEEEauthorblockA{\textit{School of Electrical and Computer Engineering} \\
Cornell University, Ithaca, NY, USA}\\
\and
\IEEEauthorblockN{Xinyi Wang}\\
\vspace{-1em}
\IEEEauthorblockA{
Menlo Park, Santa Clara, CA, USA}\\
}

\maketitle

\vspace{-2em}
\begin{abstract}
This paper introduces an Artificial Intelligence (AI) foundation model for time series in engineering applications, where causal operations are required for real-time monitoring and control. Since engineering time series are governed by physical, rather than linguistic, laws, large-language-model-based AI foundation models may be ineffective or inefficient. Building on the classical innovations representation theory of Wiener, Kallianpur, and Rosenblatt, we propose Time Series GPT (TS-GPT)---an innovations-representation-based Generative Pre-trained Transformer for engineering monitoring and control. As an example of foundation model adaptation, we consider Probabilistic Generative Forecasting, which produces future time series samples from conditional probability distributions given past realizations. We demonstrate the effectiveness of TS-GPT in forecasting real-time locational marginal prices using historical data from U.S. independent system operators. 
\end{abstract}


\vspace{1em}
\section{Introduction} \label{sec:intro}
\subsection{The Rise of AI Foundation Model}
At the heart of the current AI revolution is the Foundation Model (FM), designed to overcome computational, data, and learning challenges of AI applications across a broad range of applications. Today, FMs power some of the most successful AI applications, such as ChatGPT, demonstrating high levels of comprehension, fluency in natural language, and impressive capabilities in information extraction, synthesis, and reasoning. 

A defining feature of the architectural design of an FM is the partition of highly complex AI tasks into two processes: (i) {\em FM pretraining}, which generates critical latent features---a summary statistic---applicable across a wide range of tasks, and (ii) {\em FM adaptation,} which fine-tunes the model for specific and task-oriented applications. The genius of this partition lies in the ``division of labor'' concept of Adam Smith \cite{Smith:76}, making FM an attractive business model for AI technology. By separating the costly AI pretraining that requires vast amounts of data and immense computation power from the more application-specific FM adaptations, the FM architecture allows AI companies with AI expertise, computation resources, and financial strength to monetize FMs as commercial products from which experienced practitioners can specialize and adapt these models to meet application needs, maximizing both efficiency and applicability. 

\begin{figure}[h]
\centering
\includegraphics[width=0.7\textwidth]{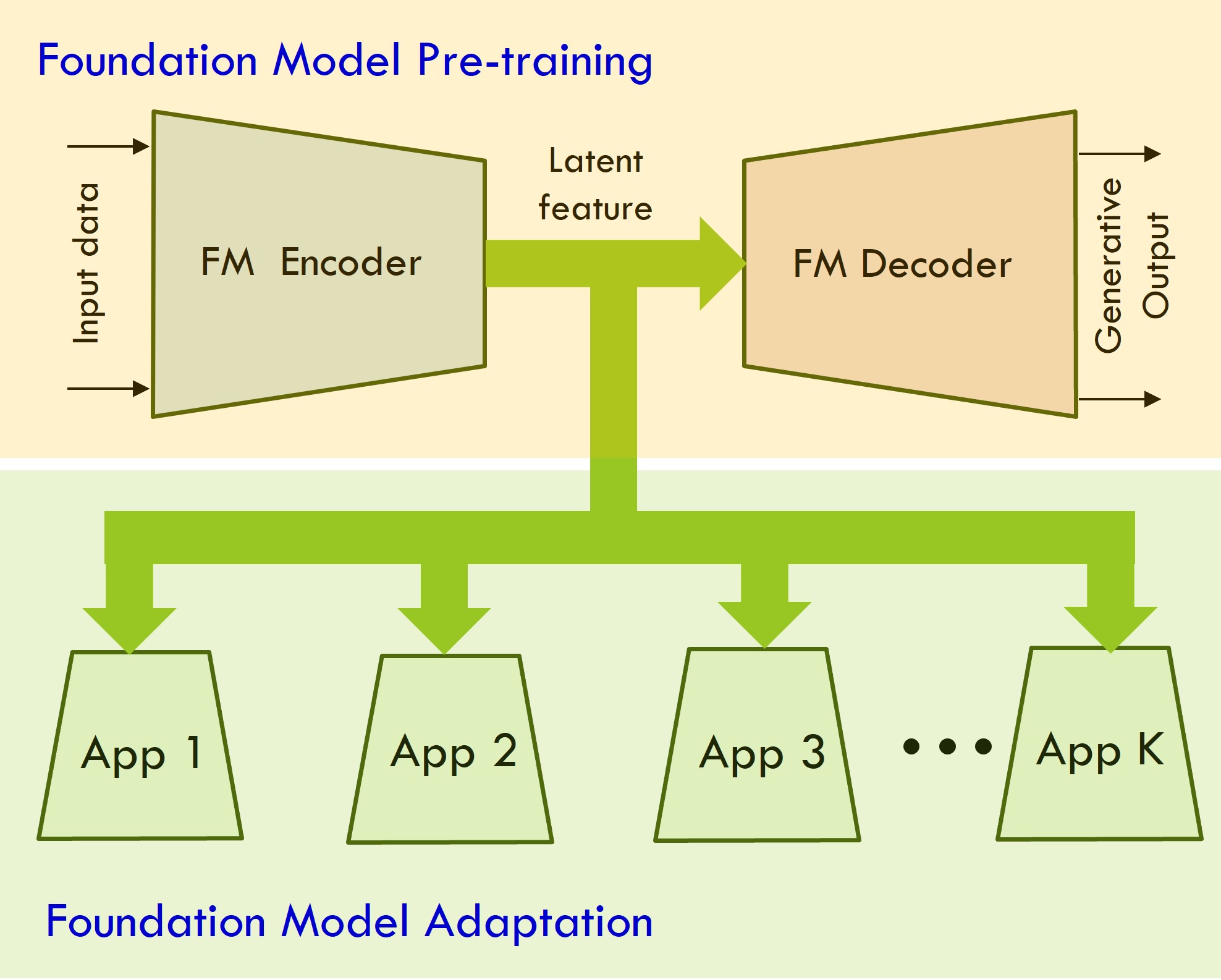}
\caption{An AI Foundation Model.}\label{fig:FM}
\end{figure}

At a high level, the architecture of FM pretraining can be abstracted as an autoencoder shown in the upper layer of Fig~\ref{fig:FM}. The encoder transforms the input into a latent representation of the input, and the decoder generates outputs that match the input according to some similarity measure. Trained with large datasets, FM is capable of producing autoencoder output that approximate input with the right underlying probabilistic structure: sentences with proper grammar and images with authentic appearance.  The FM autoencoder is also an abstraction of the so-called {\em transformer} architecture \cite{Vaswani&etal:17NIPS} where a specific ``attention mechanisms'' are embedded in the encoder-decoder  structure to capture temporal dependencies and other data characteristics.

A key feature of the FM autoencoder is its randomization that makes the FM {\em generative}, capable of producing outputs outside the training samples used in the pretraining process. The generative feature of the FM plays a crucial role in enabling FM to evolve, to acquire new capabilities as it adapts to new environments, and to learn from new data. 

Generative Pretrained Transformer, abbreviated as GPT, is the hallmark of modern AI architectures. Once pretrained, the underlying temporal dependencies of the input process are captured by the FM encoder neural network. The output of the FM encoder is the latent features representing the input, typically of lower dimensions, that can be used for various purposes: classification, prediction, language translation, interactive question-and-answer, and producing summary texts according to prompts.

Most prominent Foundation Models (FMs) are based on large language models (LLMs) \cite{Vaswani&etal:17NIPS, Bommasani&etal:22Rpt}, which are trained and optimized for tasks involving text, speech, and image/video data. While there have been efforts to apply LLM-based FMs to physical systems, it remains an open question whether these language models can effectively address real-time {\em causal} decision-making problems that rely on real-time data from systems governed by physical laws—the Ohm's, Kirchhoff's, and Maxwell's laws—other than the linguistic structures central to commercial FMs.   The lack of a mathematical foundation and interpretability are critical barriers to FM adoption in many engineering fields, particularly for critical infrastructures and control systems where considerations of physical/cyber security, operational stability, and safety are paramount.

This article focuses on FMs for time series arising from physical systems---a causal time series GPT with a canonical autoencoder structure based on an innovations representation model---aimed at developing a time series GPT approach and an adaptation for generative probabilistic forecasting. 

\vspace{1em}
\section{Innovations Through the Lens of Modern Machine Learning}

\subsection{The Wiener-Kallianpur Conjecture}
In 1958, Norbert Wiener and Gopinath Kallianpur considered the problem of efficient representations of stationary random processes \cite{Wiener:58Book}. They postulated that a stationary random process ${\bm x}:=(x_t)$ could be transformed by a {\em causal encoder} $G$ to an independent and identically distributed random sequence $\vbf(v_t)$ with the uniformly distributed marginals on $[0,1]$, herein referred to as IID-uniform. Furthermore, there exists a {\em causal decoder} $H$ that maps the sequence $\vbf$ back to the input sequence ${\bm x}$. 

Specifically, there exist causal $G$ and $H$ such that
\begin{equation} \label{eq:SIR}
\left\{\begin{array}{lll}
v_t &=G(x_t,x_{t-1},\cdots),~~&v_t\stackrel{\mbox{\sf\scriptsize IID}}{\sim}\Uc(0,1)\\
\hat{x}_t &=H(v_t,v_{t-1},\cdots),~~&\hat{{\bm x}} \stackrel{\mbox{\sf\scriptsize a.s.}}{=} {\bm x}.\\
\end{array}\right.
\end{equation}
The IID-uniform property of $\vbf$ and the almost-sure matching of ${\bm x}$ and $\hat{{\bm x}}$ imply that $v_t$ is statistically independent of the past $\xbf_{0:t-1}:=(x_{0},x_{1},\cdots, x_{t-1})$, thus the interpretation that $v_t$ represents the new information at $t$---{\em the innovation}---and $\vbf$ the {\em innovations sequence} of ${\bm x}$ \cite{Wiener&Masani:57AM-I,Wiener&Masani:57AM-II,Masani:66BAMS}.

\vspace{1em}
    \begin{figure}[h]
    \centering
    \includegraphics[width=0.9\linewidth]{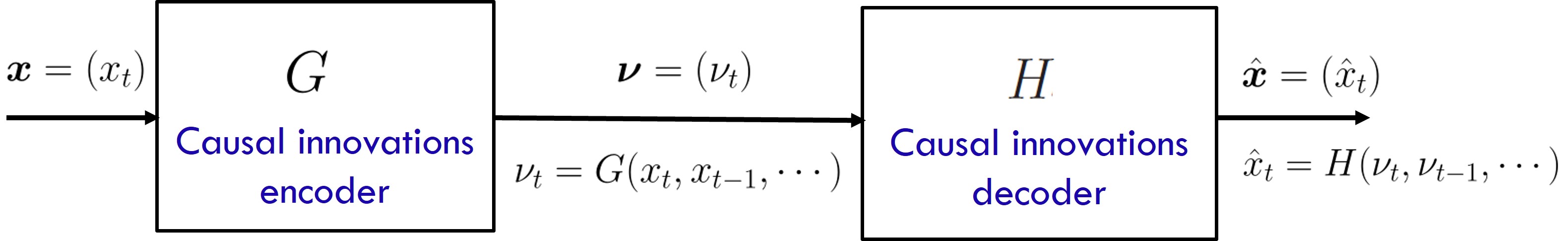}
    \caption{Innovations autoencoder with causal encoder $G$ and decoder $H$.}
    \label{fig:IAE}
\end{figure}

Through the lens of modern machine learning, the Wiener-Kallianpur conjecture suggests a universal and {\em causal} autoencoder architecture to represent stationary time series, as illustrated in Figure~\ref{fig:IAE}. The causality of the autoencoder makes the representation particularly suitable for real-time decision making. 

We call the above autoencoder the {\em Wiener-Kallianpur Autoencoder} and (\ref{eq:SIR})  the {\em Strong Innovations Representation\footnote{The word ``strong'' emphasizes that the output of the decoder matches the input on the realization basis. The innovation representation is {\em weak} if the matching is only in distribution.}} (SIR). If $(G,H)$ exists, SIR is universal, nonparametric, and powerful. The Wiener-Kallianpur autoencoder ``disentangles'' complex time series models into the autoencoder $(G,H)$ that captures the structural temporal dependencies of ${\bm x}$ and the latent sequence $\vbf$ that carries the incremental information of the input realizations. SIR  is lossless, making the innovations sequence $\vbf$ a sufficient statistic for online decision-making; the optimal decision based on ${\bm x}$ is equivalent to that based on the much simpler IID-uniform innovations $\vbf$.

Instances of innovations representations were studied before the Wiener-Kallianpur conjecture, starting from the work of Kolmogorov \cite{Kolmogorov:41}, Wiener \cite{Wiener:49Book}, Bode and Shannon \cite{Bode&Shannon:50IRE}, Wiener and Masani \cite{Wiener&Masani:57AM-I}. In a sequence of contributions \cite{Kailath:68TAC_I,Kailath&Frost:68TAC_II,Geesey&Kailath:69,Kailath:69_II}, Kailath and his co-authors made significant contributions that popularized the idea of innovations representation in engineering fields, particularly in control, communications, and signal processing.  
See \cite{Kailath:70,Kailath:74} for surveys of the field at the time.

For practical applications, SIR is the most powerful when the autoencoder  $(G,H)$ exists and can be computed. Most of the known cases in which a causal autoencoder can be obtained explicitly or from data involve Gaussian assumptions. The simplest case is the stationary Gaussian process, for which $(G,H)$ can be computed from the (linear) minimum-mean-squared-error (MMSE) prediction error filter. By the orthogonality principle, the prediction error at every time is statistically independent of the past, making the prediction errors a sequence of innovations. 

Kailath and Frost made a key contribution to the innovations representation of a particular type of non-Gaussian continuous-time random processes \cite{Kailath:71,Frost&Kailath:71TAC_III}. In particular, they considered the class of additive white Gaussian noise (AWGN) processes defined as the sum of a (possibly) non-Gaussian stationary process and a white Gaussian noise (Wiener) process. Under mild assumptions that hold favorably in practical scenarios, they showed the remarkable result parallel to the Gaussian case: the innovations process is the output of the  MMSE ({\em nonlinear}) prediction error filter. The Kailath-Frost innovations representation is especially appealing because it not only generalizes the SIR of Gaussian processes but also provides a method to extract the innovations process. Unfortunately, the SIR for the continuous-time AWGN processes does not translate directly to the discrete-time AWGN process. However, it is arguable that the nonlinear MMSE prediction error sequence can serve as a good approximation of the actual innovations sequence in practice. Indeed, the training of a time series foundation model with strong innovation represents developed in Sec.~\ref{sec:TimeSeriesGPT} follows the idea of nonlinear MMSE prediction.   

\subsection{Weak Innovations Representation}
Powerful as SIR is, the generality of SIR was brought into question by Murray Rosenblatt, who constructed counterexamples for which SIR does not exist ``even in the case of a finite-state purely nondeterministic Markov chain''  \cite{Rosenblatt:59,Rosenblatt:09}. Although the existence of a causal mapping that extracts an IID-uniformly distributed random sequence is easily satisfied under mild assumptions, establishing the existence of causally invertible mappings is challenging. It remains an open problem to characterize the general existence conditions of SIR \cite{Rosenblatt:09,Wu:05PNAS,Wu:11}.

However, Rosenblatt considered a relaxation of the perfect reconstruction condition in (\ref{eq:SIR}) to the weaker version that requires the input and output of the autoencoder to match only in distribution\footnote{Rosenblatt credited Paul Levy for suggesting this relaxation in \cite{levy:37book}}: 
\begin{equation} \label{eq:WIR}
\left\{\begin{array}{lll}
v_t &=G(x_t,x_{t-1},\cdots),~~&v_t\stackrel{\mbox{\sf\scriptsize IID}}{\sim}\Uc(0,1)\\
\hat{x}_t &=H(v_t,v_{t-1},\cdots),~~&\hat{{\bm x}} \stackrel{\mbox{\sf\scriptsize d}}{=} {\bm x}.\\
\end{array}\right.
\end{equation}
We shall call the representation (\ref{eq:WIR}) {\em weak innovations representation (WIR)}, $\vbf$ the {\em weak innovations sequence}, and the autoencoder pair $(G,H)$ a {\em Weak Innovations AutoEncoder (WIAE)}. Rosenblatt showed that WIR exists for the purely nondeterministic finite-state stationary Markov chain---the cases in which SIR does not exist except for rare and nearly pathological scenarios.   

The benefit of WIR for being applicable to broader classes of random processes comes with nontrivial costs. First, without requiring perfect reconstruction of the autoencoder input at its output, WIR representation cannot be used for certain applications, such as compression, when accurate recovery of the source is the objective. 

Second, while a strong innovations sequence is a {\em sufficient statistic}, a weak innovations sequence is not in general. Therefore, decision-making based on weak innovations may be suboptimal. There are important exceptions, however. For the probabilistic forecasting problem discussed in Sec.~\ref {sec:GPF}, we show that the weak innovations sequence defined by (\ref{eq:WIR}) is {\em Bayesian sufficient} and that using the weak innovations sequence for probabilistic forecasting is optimal.

Finally, it is significant that the weak innovations sequence $\vbf$  does not have the same interpretation that $v_t$ is statistically independent of the past $\Xmsc_{t-1}:=(x_{t-1},x_{t-2},\cdots)$ as required by SIR (\ref{eq:SIR}). Instead, $v_t$ is statistically independent of the past autoencoder {\em output} $\hat{\Xmsc}_{t-1}:=(\hat{x}_{t-1},\hat{x}_{t-2},\cdots)$.  In other words, the weak innovations are the (strong) innovations of $\hat{{\bm x}}$ rather than ${{\bm x}}$.

\vspace{1em}
\section{A Time Series GPT Foundation Model}\label{sec:TimeSeriesGPT}
The strong and weak innovations representations, initially envisioned by Wiener, Kallianpur, and Rosenblatt,  give perhaps the most succinct {\em causal} representations of general stationary processes with the autoencoder $(G,H)$ capturing the statistical temporal dependencies and its latent process $\vbf$ the realized randomness of the underlying random process. Embedding a strong (or weak) innovations representation as the information processing engine of an AI foundation model for real-time decision-making is natural. 

This section focuses on the Generative and Pretrained Transformer FM architecture for time series, illustrated in Fig.~\ref{fig:FM}. Specifically, the pretraining of the FM involves using historical data to train the autoencoder $(G,H)$ to extract the innovations sequence $\vbf$ as the latent feature of the realized input $\xbf$.  The autoencoder may include attention mechanisms, not necessarily the same in the same form of as the standard query-key-value implementations, that focus the attention to specific characteristics of the data such as spikiness and long range dependencies.  

The generative feature of the time series FM is implemented by the decoder $H$. By replacing the innovations sequence  $\vbf$  at the encoder output with independently generated {\em pseudo-innovations} $\tilde{\vbf}$ from IID-uniform distributions, the autodecoder $H$  produces out-of-sample data $\tilde{\xbf}$ with the same distribution as but different realizations from the training data population.

\subsection{Pretraining Foundation Model with Innovations Representation}

Except for the Gaussian and AWGN (continuous-time) processes, there had not been computationally tractable solutions to construct strong or weak innovations autoencoders, severely limiting the applications of innovations representations. Here, we describe, at the architecture-level, the learning from data an autoencoder pair $(G,H)$ for the strong or weak innovations representation, based on a variation of the generative adversarial networks (GAN) learning strategy, following the first such development in \cite{Wang&Tong:22JMLR}.

The architecture of the learning of the strong and weak innovations representations is illustrated in Fig.~\ref{fig:IAE}, where the autoencoder $(G,H)$ is realized by deep neural networks $(G_{\theta},H_{\eta})$ with coefficients $(\theta,\eta)$, respectively. The dashed (red) and solid (black) lines are signal flow paths during pre-training.  After training, only the solid (black) lines remain.  

\vspace{1em}
    \begin{figure}[h]
    \centering
    \includegraphics[width=0.7\linewidth]{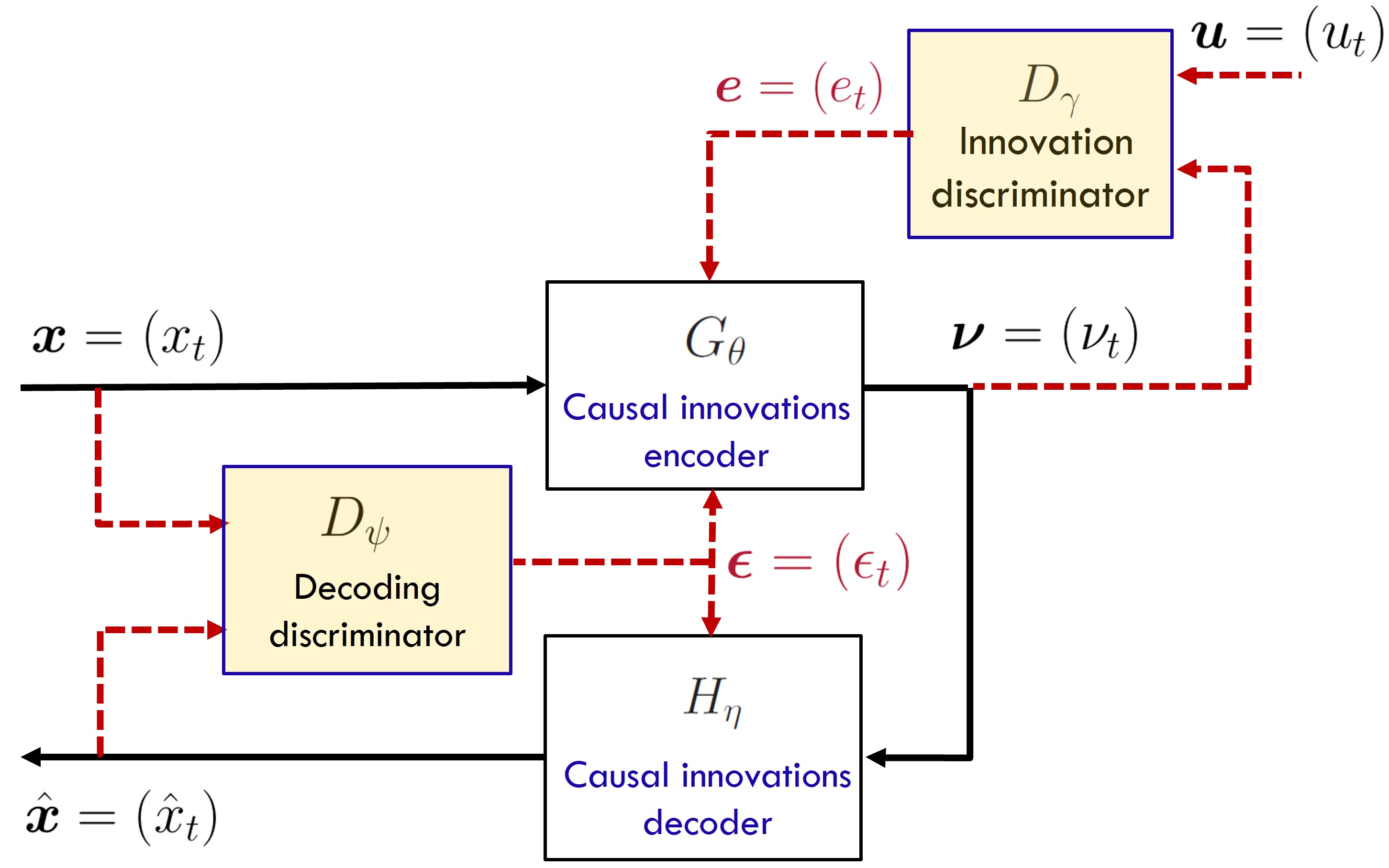}
    \caption{A deep learning architecture of innovations representations.}
    \label{fig:IAE}
\end{figure}

At time $t$, the input of the causal encoder $G_\theta$  consists of the current and past samples $\xbf_t:=(x_t,x_{t-1},\cdots)$, and the output of $G$ is the innovation sequence $\vbf_t:=(v_t,v_{t-1},\cdots)$.
 Likewise, the input of the decoder $H_\eta$  is the innovation sequence  $(\vbf_t)$, and the output $(\hat{\xbf}_t)$ the estimate of $\xbf_t$.

The objective of training the autoencoder pair $(G_{\theta},H_{\eta})$ toward its optimal setting is to drive the encoder $G_{\theta}$ to output an IID-uniform sequence as required by the innovations sequence and the decoder  $H_{\eta}$ to output $\hat{\xbf}$ that matches $\xbf$, in distribution for WIR and in the mean-squared sense for SIR.  

To generate training updates for the encoder $G_\theta$, an {\em innovations discriminator} neural network $D_\gamma$ compares  the encoder output $\vbf$ with a synthetically generated IID-uniform sequence $\ubf$ and produces a Wasserstein distance error sequence $\ebf$ to update to update parameter $\theta$ of the encoder $G_\theta$. Likewise, 
the decoding discriminator $D_\psi$ compares the autoencoder input $\xbf$ and output $\hat{\xbf}$ and generates error sequence $\epsilonbf$ to updates jointly $(\theta,\eta)$ of the encoder $G_\theta$ and decoder $H_\eta$. For SIR, 
$D_\psi$ derives updates from the stochastic gradient of the mean squared error between $\xbf$ and $\hat{\xbf}$.
For WIR,  $D_\psi$ derives updates from the stochastic gradient of the Wasserstein distance error as in $D_\gamma$. The computation of Wasserstein distance error is standard. 

Define the overall pretraining objective as the weighted sum of innovations sequence error at the output of the innovations discriminator and the decoding error at the output of the decoding discriminator:
\begin{equation*}
L(\theta,\eta,\gamma,\psi) :=\mathbb{E}[D_\gamma(\boldsymbol{v},\boldsymbol{u})]
        +\lambda\mathbb{E}[D_\psi(\boldsymbol{\hat{x}},\boldsymbol{x})].
\end{equation*}
Through the Kantorovich-Rubinstein duality \cite{Villani09:Book}, the neural network parameters $(\theta,\eta,\gamma,\psi)$ are obtained from the min-max  optimization:
\begin{equation}
\min_{\theta,\eta}\max_{\gamma,\psi} L(\theta,\eta,\gamma,\psi).
    \label{eq:train}
\end{equation}
Standard stochastic gradient descent algorithms can be used to train $(D_\eta,D_\psi)$. A pseudo-code with a detailed training procedure can be found in an unpublished arXiv article \cite{Wang&Lee&Tong&Zhao:22arxiv}.   
Our experience showed that the training of WIR appears easier than that of SIR, thanks to the use of the Wasserstein GAN in the decoding discriminator.

\subsection{Structural Convergence}
The existence of SIR and WIR requires the autoencoder $(G,H)$ to take input samples from the infinite past. While a recurrent neural network autoencoder could be used, the training of such an autoencoder is challenging. If the autoencoder $(G_\theta,H_\eta)$ and related discriminators $(D_\gamma,D_\psi)$ are implemented with large but finite-dimensional feedforward convolutional neural networks that input from only the finite past, can the learned autoencoder approximate the true strong/weak innovations well? 

In \cite{Wang&Tong:22JMLR}, structural convergence is established for strong innovations representations. The same can be shown for the weak representations. Assume that all neural networks in Fig.~\ref{fig:IAE} take as their input the  past $k$ data samples. Let the finite $k+1$-input dimensional autoencoder be $(G^{(k)}_\theta,H^{(k)}_\eta)$, and
\begin{equation} \label{eq:WIR-k}
\left\{\begin{array}{ll}
v_t^{(k)} &:=H^{(k)}(x_t,x_{t-1},\cdots, x_{t-k}),\\
\hat{x}^{(k)}_t &:=G^{(k)}\Big(v^{(k)}_t,v^{(k)}_{t-1},\cdots,v^{(k)}_{t-k}\Big).
\end{array}\right.
\end{equation}
Let the finite input-dimensional discriminators $(G^{(k)}_\theta,H^{(k)}_\eta)$ be similarly defined. It can be shown that, as $k\rightarrow \infty$, $\vbf^{(k)} \stackrel{\scriptsize{\sf m.s.}}{\rightarrow} \vbf$ and  $\xbf^{(k)} \stackrel{\scriptsize{\sf m.s.}}{\rightarrow} \xbf$ in the mean-squared sense, assuming that the training of $(G^{(k)}_\theta,H^{(k)}_\eta)$ based on (\ref{eq:train}) converges globally for every $k$.

\section{Probabilistic Forecasting GPT with Innovations Representation} \label{sec:GPF}
With the time series GPT discussed in the previous section, this section focuses on generative probabilistic forecasting of the time series at future instances. For notational simplicity, we consider only the prediction of a scalar random process $(X_t)$. Here, a capital letter $X$ denotes a random variable and its lower case $x$ the realization.

\subsection{Generative Probabilistic Forecasting}
Given observation $\xbf_{0:t}=(x_0,\cdots, x_t)$ of a time series $(X_t)$ up to time $t$, the classical $T$-step-ahead {\em point forecasting} is to produce an estimate $\hat{x}_{t+T}=f(\xbf_{0:t})$ of $x_{t+T}$.  A conventional approach is to find the estimator $f$ that minimizes the conditional (and unconditional) mean-squared or mean absolute error.

{\em Probabilistic Forecasting}, in contrast, is to obtain an estimate $\hat{F}_{t+T|t}$ of the conditional distribution of 
\begin{equation*}
F_{t+T|t}(x|\xbf_{0:t}):=\Pr(X_{t+T}\le x|\Xbf_{0:t}=\xbf_{0:t}).
\end{equation*}
Once the conditional distribution $F_{t+T|t}$ is estimated, point estimates can be computed, with the minimum mean-squared-error (MMSE) being the conditional mean and the minimum mean-absolute-error (MMAE) the conditional median. 

Probabilistic forecasting is exceedingly difficult without parameterizing the distribution; standard techniques of nonparametric distribution estimation cannot be applied, because $F_{t+T|t}$ is a function of the past $\xbf_{0:t}$. Given the trajectory  $\xbf_{0:t}$ of the time series up to time $t$, there is a single realization $x_{t+T}$ associated with the observed history $\xbf_{0:t}$, making it intractable to estimate $F_{t+T|t}$. A standard approach is to assume that the time series can be parameterized by a finite-dimensional parameter $\thetabf$, say, the mean and variance of a stationary Gaussian process.

{\em Generative Probabilistic Forecasting (GPF)} is to produce realizations with the conditional distribution $ F_{t_{t+T}|t}$. Instead of producing a single ``best estimate'' of the realized $X_{t+T}$, a GPF is to produce an ensemble of realizations of $X_{t+T}$ given the past data $\xbf_{0:t}$. In a way, GPF and probabilistic forecasting are estimating the same object---the conditional distribution $F_{t+T|t}$. Having obtained a probabilistic forecast $\hat{F}_{t+T|t}$, one can produce an ensemble of realizations via Monte Carlo simulations. On the other hand, having obtained a GPF, one can produce a distribution estimate by standard nonparametric techniques by applying the law of large numbers. A key difference between probabilistic forecasting and GPF is that, as we show next, obtaining a GPF from data is considerably simpler. 

\subsection{Structure of a Generative Probabilistic Forecaster}
\begin{figure}[h]
    \centering
    \includegraphics[width=0.8\linewidth]{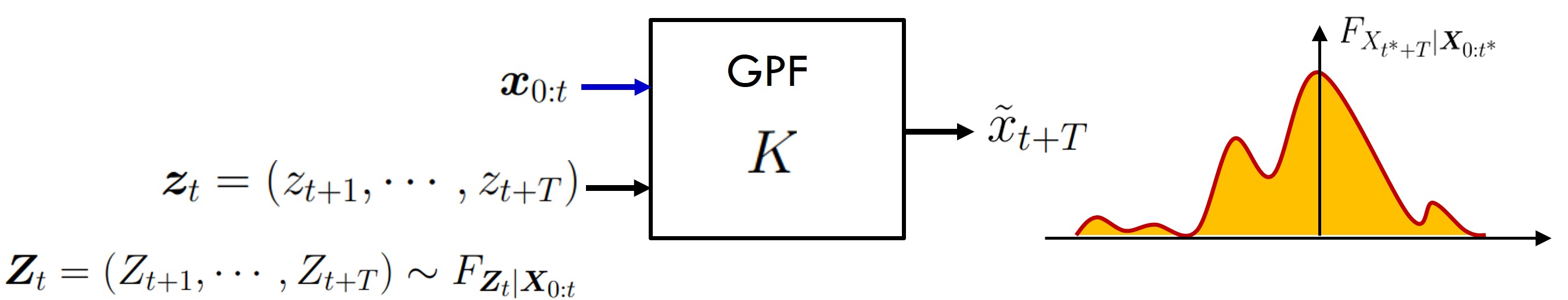}
    \caption{Structure of Generative Probabilistic Forecaster (GPF).}
    \label{fig:GPF}
\end{figure}

As shown in Fig.~\ref{fig:GPF}, given $\Xbf_{0:t}=\xbf_{0:t}$, a generative probabilistic forecaster (GPF) is a randomized mapping $K$ that maps past observations $\xbf_{0:t}$ and a randomization vector $\Zbf_t=(Z_{t+1},\cdots, Z_{t+T})$ to random variable $\tilde{X}_{t+T}$ having the same conditional distribution as $X_{t+T}$, \ie 
\begin{equation*}
\tilde{X}_{t+T} = K(\xbf_{0:t},\Zbf_t) \sim F_{t+T|t},
\end{equation*}
where the randomization vector $\Zbf_t=(Z_{t+1},\cdots, Z_{t+T})$ follows some sampling distribution $F_{\Zbf_t|\Xbf_{0:t}}$. Thus, the design of GPF is to find a mapping $K$ and a sampling distribution $F_{\Zbf_t|\Xbf_{0:t}}$.

\vspace{1em}

\begin{figure}[h]
    \centering
    \includegraphics[width=0.8\linewidth]{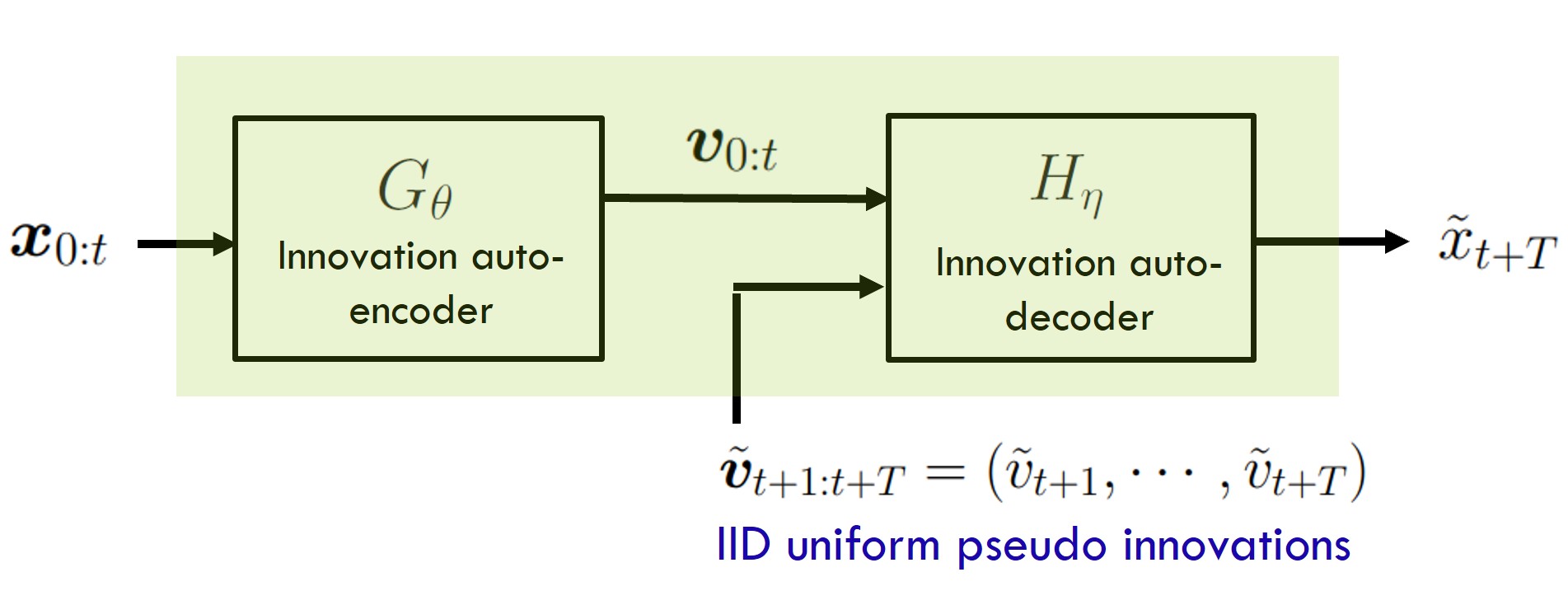}
    \caption{Generative Probabilistic Forecaster (GPF).}
    \label{fig:GPF2}
\end{figure}

A trivial but impractical GPF is to choose the sampling probability distribution directly as $F_{X_{t+T}|\Xbf_{0:t}}$, which is unknown, unfortunately. On the other hand, the innovations representation model immediately suggests a generative probabilistic forecaster, thanks to the IID-uniform nature of the innovations sequence that decouples the future from the current and the past. The lack of future realizations of the innovation sequence can be replaced by independently generated pseudo-innovations that are IID-uniform, as shown in Fig.~\ref{fig:GPF2}. 
Specifically, given the realized time series $\xbf_{0:t}$ up to time $t$, the generative forecast $\tilde{x}_{t+T}$ of $x_{t+T}$ is given by
\begin{equation}
\tilde{x}_{t+T}=H_\eta(\vbf_{0:T},\tilde{\vbf}_{t+1:t+T}), \label{eq:WIAE}
\end{equation}
where $\vbf_{0:T}=(v_0,\cdots, v_t)$ is the innovations sequence of $\xbf_{0:t}$ and $\tilde{\vbf}_{t+1:t+T}=(v_{t+1},\cdots,v_{t+T})$ the pseudo innovations of unrealized time series $\xbf_{t:t+T}$.

The validity of the above construction can be established as follows. Here we assume the ideal training of the weak innovation autoencoder $(G_\theta,H_\eta)$, such that the latent innovations sequence $\Vbf=(V_t)$ is IID-uniform, and the input and output of the autoencoder match in distribution. In particular have
\begin{equation*}
\Pr\big[X_{t+T}\le x\big|\Xbf_{0:t}=\xbf_{0:t}\big] =
\Pr\big[\hat{X}_{t+T}\le x\big|\Xbf_{0:t}=\xbf_{0:t}\big]
\end{equation*}
Assume that the ideally trained autoencoder $G_\theta$ is injective---a theoretically limiting but practically reasonable for deep neural network representations. Let $\tilde{\Vbf}_t=(\tilde{V}_{t+1},\cdots, \tilde{V}_{t+T})$ be the peseudo-innovations vector independent of the weak innovations $\Vbf=(V_t)$ and
$\tilde{X}_{t+T}=H(\Vbf_{0:t},\tilde{\Vbf}_t)$ the output of the auto-decoder with innovations vector $\Vbf_{t+1:t+T}$ replaced by $\tilde{\Vbf}_t$.  Then,
\begin{align}
\Pr\big[\hat{X}_{t+T}\le x\big|\Xbf_{0:t}=\xbf_{0:t}\big] & =
\Pr\big[\hat{X}_{t+T}\le x\big|\Vbf_{0:t}=\vbf_{0:t}\big] \label{eq:suff}\\
&=\Pr\big[H_\eta(\vbf_{0:t},\Vbf_{t+1:t+T})\le x\big|\Vbf_{0:t}=\vbf_{0:t}\big]\nonumber\\
&=\Pr\big[H_\eta(\vbf_{0:t},\tilde{\Vbf}_{t+1:t+T})\le x\big|\Vbf_{0:t}=\vbf_{0:t}\big]\nonumber\\
&=\Pr\big[\tilde{X}_{t+T}\le x\big|\Vbf_{0:t}=\vbf_{0:t}\big], \label{eq:GPFvalid}
\end{align}
where we used the fact that $(\Vbf_{0:t},\Vbf_{t+1:t+T},\tilde{\Vbf}_t)$ are jointly independent and $(\Vbf_{0:t},\Vbf_{t+1:t+T})\stackrel{\mbox{\scriptsize d}}{=}(\Vbf_{0:t},\tilde{\Vbf}_t)$. 

A few remarks on the above derivation are in order.
\begin{itemize}
\item {\bf Bayesian Sufficency:}  Equation (\ref{eq:GPFvalid}) implies that the weak innovation sequence is {\em Bayesian sufficient} for decision involving any random variable $X_{t+T}$ at a future time, in the sense that conditioning on innovations up to time $t$ incurrs no loss comparing to conditioning on the raw data $\Xbf_{0:t}$.

\item {\bf Validity of GPT:}   With (\ref{eq:GPFvalid}), we show that the GPF shown in Fig.~\ref{fig:GPF2} gives samples with the correct conditional probability distribution, under the ideal pretraining of the innovations autoencoder. Note that the above GPF can easily generate an arbitrarily large number of independent samples, from which conditional probability distributions can be estimated using standard distribution estimation techniques.
\end{itemize}

\subsection{From GPF to Point and Quantile Forecasting}
GPF can produce an arbitrarily large number of independently generated samples of the conditional probability distribution, from which point and quantile forecasts can be readily computed. 

Let  $\left\{\tilde{x}_{t}^{(k)}, k=1,\cdots, K\right\}$ be the set of $K$ independently generated samples (by a GPF) following the conditional probability distribution $F_{t+T|t}$ of the time series for $X_{t+T}$ given past observations $\xbf_{0:t}$ up to time $t$. For the simplicity of mathematical expressions, we assume that $\left\{\tilde{\xbf}_{t}^{(k)}\right\}$ is sorted in an ascending order.

Some of the most popular point estimates are computed as follows:
\begin{itemize}
\item  {\bf Minimum Mean-Squared-Error (MMSE) Forecast:} The MMSE forecast is the mean of the conditional distribution. The MMSE forecast $\hat{\xbf}^{\mbox{\scriptsize\sf  MMSE}}_{t}$  by a GPF is given by the conditional sample mean
\begin{equation*}
\hat{\xbf}^{\mbox{\scriptsize\sf  MMSE}}_{t} =\frac{1}{K} \sum_{k=1}^K \tilde{\xbf}_{t}^{(k)}.
\end{equation*}
\item  {\bf Minimum Mean-Absolute-Error (MMAE) Forecast:} The MMAE forecast is the median of the conditional distribution.   The MMAE forecast $\hat{\xbf}^{\mbox{\scriptsize\sf  MMAE}}_{t}$  by a GPF is given by the conditional sample median
\begin{equation*}
\hat{\xbf}^{\mbox{\scriptsize\sf MMAE}}_{t} = \begin{cases}
    \tilde{\xbf}_{t}^{\left((K+1)/2\right)}, &\mbox{if $K$ is odd}\\
    0.5\left(\tilde{\xbf}_{t}^{\left(K/2\right)}+\tilde{\xbf}_{t}^{\left(K/2+1\right)}\right), &\mbox{if $K$ is even}.\\
\end{cases}
\end{equation*}
\item {\bf Quantile-Forecast:}  The forecast of $q$-quantile $\hat{\xbf}^{\mbox{\scriptsize\sf  $q$-QT}}_{t}$ is given by:
\begin{equation*}
\hat{\xbf}^{\mbox{\scriptsize\sf $q$-QT}}_{t} = \begin{cases}
    \tilde{\xbf}_{t}^{\left(qK\right)}, &\mbox{if $qK$ is an integer}\\
0.5\left(\tilde{\xbf}_{t}^{\left([qK]\right)}+\tilde{\xbf}_{t}^{\left([qK]+1\right)}\right), &\mbox{otherwise},\\
\end{cases}
\end{equation*}
where $[a]$ indicates the greatest integer not exceeding $a$.
\end{itemize}

\subsection{Performance Measure: CRPS, CPE and ACPE}
Evaluating the performance of probabilistic forecasting and GPF is nontrivial. The difficulty lies in that the ground truth of the forecast quantity---the conditional probability distribution $F_{t+T|t}$ of the future random variable $X_{t+T}$ given the past $\xbf_{0:t}$---is unknown. 

\paragraph{Continuous Ranked Probability Score (CRPS)} 
A commonly used probabilistic forecasting metric is the
{\em Continuous Ranked Probability Score (CRPS).} Let $\hat{F}_{t+T|t}$ be the estimated conditional CDF of $X_{t+T}$ given $\Xbf_{0:t}=\xbf_{0:t}$ and $x_{t+T}$ the realized $X_{t+T}$. CRPS and the expected CRPS are defined by
\begin{align*}
    \mbox{\sf CRPS}\big(\hat{F}_{t+T|t},x_{t+T}\big)&:=\int_{-\infty}^\infty \Big(\hat{F}_{t+T|t}(z|\xbf_{0:t})-\mathbbm{I}\{x_{t+T}\le z\}\Big)^2 dz\\
    \overline{\mbox{\sf CRPS}}(\hat{F}_{t+T|t})&:=\mbbE\Big[\mbox{\sf CRPS}(\hat{F}_{t+T|t},X_{t+T})\Big], 
\end{align*}
where $\mathbbm{I}$ is the indicator function.  It can be easily shown that $\overline{\mbox{\sf CRPS}}(\hat{F}_{t+T|t})$ is minimum (albeit not necessarily zero) when $\hat{F}_{t+T|t}=F_{t+T|t}$.  

In practice, given a realization of the time series $(X_t=x_t)$ of length $N$, we approximate $\overline{\mbox{\sf CRPS}}(\hat{F}_{t+T|t})$ by the empirical sum
\begin{equation*}
\widehat{\mbox{\sf CRPS}}
= \frac{1}{N-T} \sum_{t=T-1}^N \int_{-\infty}^\infty 
\Big(\hat{F}_{t|t-T}(z|\xbf_{0:t-T})-\mathbbm{I}\{x_{t}\le z\}\Big)^2dz.
\end{equation*}
For generative probability forecasting that produces only generative forecast samples $\hat{\Xc}:=\{\hat{x}_k\}$, $\hat{F}_{t+T|t}$ is estimated from $\hat{\Xc}$.

\paragraph{Coverage Probability Error (CPE)} While CRPS measures the accuracy of a GPF over the domain of $F_{t+T|t}$, the  {\em Coverage Probability Error} at level $\alpha$, denoted by $\mbox{\sf CPE}_\alpha$, evaluates  the accuracy of the predicted $\alpha\times 100 \%$ coverage interval\footnote{Typically, the $\alpha$ coverage interval is the symmetrical interval with respect to the mean with probability $\alpha$.} based on $\hat{F}_{t+T|t}$. 

Let $[L^\alpha_{t+T},U^\alpha_{t+T}]$ be the (random conditional) $\alpha$-coverage interval of $X_{t+T}$ under the ground truth conditional distribution $F_{t+T|t}$, \ie
\begin{equation*}
\Pr\Big(X_{t+T} \in  [L^\alpha_{t+T},U^\alpha_{t+T}]\Big| \Xbf_{0:t}=\xbf_{0:t}\Big)=\alpha.
\end{equation*}
Let $[\hat{L}^\alpha_{t+T},\hat{U}^\alpha_{t+T}]$ be the {\em predicted} conditional $\alpha$ coverage interval of $X_{t+T}$ under the predicted conditional distribution $\hat{F}_{t+T|t}$.  The {\em Coverage Probability Error (CPE)} is defined by
\begin{equation*}
\mbox{\sf CPE}_\alpha(\hat{F}_{t+T|t}):=\mbbE\Big[\Pr\Big(X_{t+T} \in  [\hat{L}^\alpha_{t+T},\hat{U}^\alpha_{t+T}]\Big| \Xbf_{0:t}\Big)-\alpha\Big],
\end{equation*}
which is zero if $\hat{F}_{t+T|t}=F_{t+T|t}$.  The {\em Absolute CPE} at the coverage level $\alpha$ is defined by
\begin{equation*}
\mbox{\sf ACPE}_\alpha(\hat{F}_{t+T|t}) :=\Big|\mbox{\sf CPE}_\alpha(\hat{F}_{t+T|t})\Big|.
\end{equation*}

Given the realizations of the time series and the sequence of predicted $\alpha$-coverage intervals $\{[\hat{l}^\alpha_{t+T},\hat{u}^\alpha_{t+T}], t=0:N-T\}$, the CPE and ACPE of a GPF are approximated by sample averages.  In particular, the estimated ACPE $\widehat{\mbox{\sf ACPE}}_\alpha$ is computed by
\begin{equation*}
\widehat{\mbox{\sf ACPE}}_\alpha:=\Bigg|\frac{1}{N-T}\sum_{t=T}^{N}\mathbbm{1}\{x_{t} \in [\hat{l}^\alpha_{t-T},\hat{u}^\alpha_{t-T}]\}-\alpha\Bigg|,
\end{equation*}
where the predicted coverage bounds $(\hat{l}^\alpha_{t-T},\hat{u}^\alpha_{t-T})$ are computed from the generated samples $\big\{\tilde{x}^{(k)}_{t+T}, k=1,\cdots, K\big\}$ by the generative forecaster.

\vspace{1em}
\section{Application: GPF of Real-time Electricity Prices}
We illustrate the application of the innovations representation model-based FM for generative probabilistic forecasting of real-time prices of electricity using published LMP data by the New York Independent System Operator (NYISO) to train the FM and evaluate to the forecasting performance. More extensive empirical study results can be found in \cite{Wang:24Thesis}.

\begin{figure}[h]
    \centering
    \includegraphics[width=0.8\linewidth]{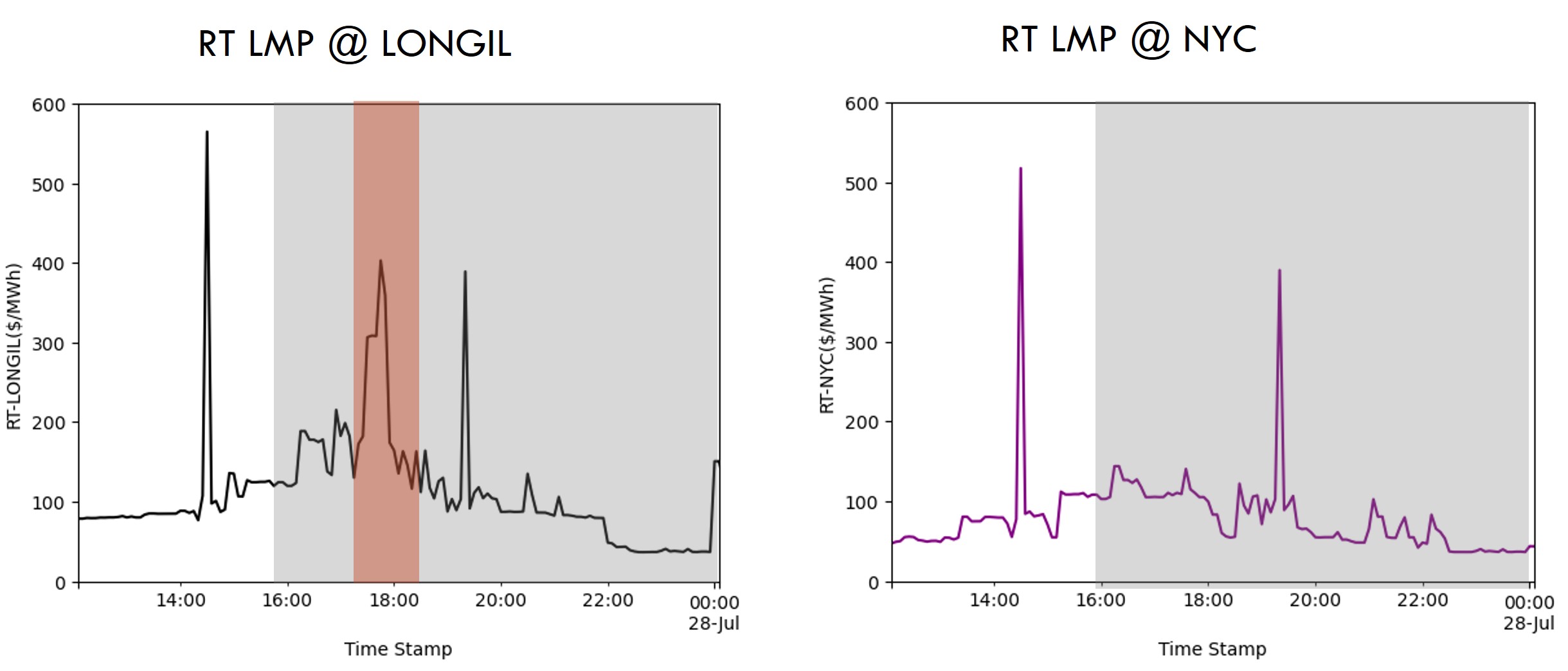}
    \caption{Locational Marginal Prices at the Long Island and New York City buses on July 26, 2027.}
    \label{fig:NYISO-LMP}
\end{figure}

\paragraph{Locational Marginal Price (LMP)}  In deregulated electricity markets in the U.S., electricity prices are referred to as {\em locational marginal prices (LMP)}, set every five minutes based on bids from demands, offers from supplies, and locations of trades.  Unlike time series of physical processes such as wind, solar, and aggregated electricity consumption, LMPs are computed from dual variables associated with power flow constraints of a convex optimization.  Fig.~\ref{fig:NYISO-LMP} shows the actual electricity prices at the Long Island (LONGIL) and New York City (NYC) buses on July 26, 2023.

\vspace{1em}

Unlike time series of physical processes such as wind, solar, and aggregated electricity consumption, LMPs can be highly volatile with huge spikes that could jump hundreds, even thousands, of dollars per megawatt-hour.  The volatility of LMP makes probabilistic price forecasting an essential risk-mitigating tool. 

\paragraph{Dataset and Forecasting Settings}  A typical deregulated real-time electricity market operates at the 5-minute timescale.  NYISO publishes decades of 5-minute real-time LMPs for all its buses.

The setup of the GPF of real-time LMP follows the participation requirement of real-time electricity market, where bids and offers must be submitted 30 to 75 minutes ahead of the market clearing. We therefore set the forecast horizon $T=12$ (60 minutes).  We assume that the FM is updated weakly using the past 30 days of LMP data.

\paragraph{Empirical Evaluations}  Extensive performance evaluations are reported in \cite{Wang:24Thesis} .  Here we present a sample of results based on the 2023-2024 LMP and demand traces, where we selected four months, one in each season (July and October in 2023, and January and April in 2024) for evaluation.  We choose the Long Island bus (LONGIL) for evaluating the probabilistic forecasting performance of 60-minute ahead forecaster based on X-minute past LMP values at LONGI and NYC and the day-ahead LMP at LONGIL and system demand.

\begin{figure}[h]
    \centering
    \includegraphics[width=0.95\linewidth]{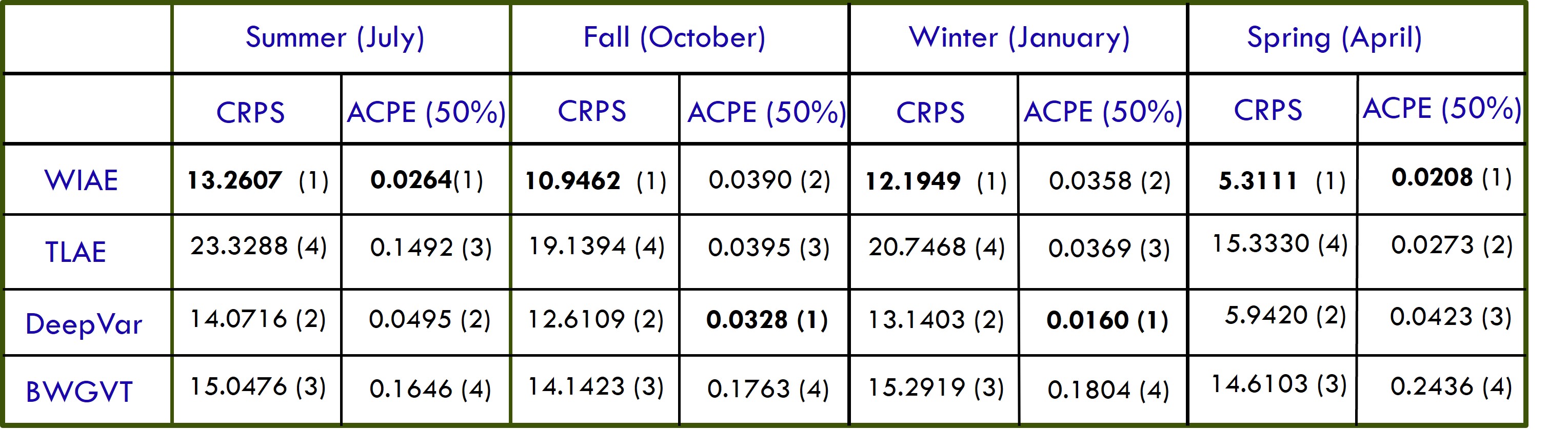}
    \caption{Performance comparisons of GPF benchmarks. Boldface numbers are the best scores. The numbers in brackets are rankings.}
    \label{fig:GPFPerformance}
\end{figure}

We compared the GPF with the weak innovations autoencoder forecaster, abbreviated by WIAE and defined in (\ref{eq:WIAE}), with several leading GPF benchmarks: 
TLAE \cite{Nguyen&Quanz:21}, DeepVAR \cite{SalinasEtal:19NeuripsDeepVAR}, and the more recent large language model (LLM) based BWGVT \cite{BottieauEtal:23TPS}. Fig.~\ref{fig:GPFPerformance} shows the 60-minute ahead probabilistic forecasting performance under CRPS and 50\%-CPE metric, where WIAE demonstrated superior performance for most test cases.

\vspace{1em}

\vspace{1em}
\section{Conclusion}
Through the lens of modern machine learning and AI foundation models, this article aims to connect modern machine learning and AI methodology with the classical innovations representation theory and powerful model-based practical solutions such as Wiener and Kalman filtering, matched filtering, and optimal control, leading to data-driven approaches that are grounded sound mathematical principles. To this end, the innovations representation model can be a powerful engine for AI foundation models for time series applications.

{
\bibliographystyle{IEEEtran}
\bibliography{LangTongBib}
}


\end{document}